\documentclass[10pt,journal,compsoc]{IEEEtran}
\ifCLASSOPTIONcompsoc
  \usepackage[nocompress]{cite}
\else
  \usepackage{cite}
\fi
\ifCLASSINFOpdf
  \usepackage[pdftex]{graphicx}
\else
\fi
\usepackage[cmex10]{amsmath}
\ifCLASSOPTIONcompsoc
 \usepackage[caption=false,font=footnotesize,labelfont=sf,textfont=sf]{subfig}
\else
 \usepackage[caption=false,font=footnotesize]{subfig}
\fi
\begin{document}
\title{Bridging Phylogeny and Taxonomy with Protein-protein Interaction Networks}

\author{Long-Huei Chen*,
        Mohana Prasad Sathya Moorthy*,
        and~Pratyaksh~Sharma*\thanks{*  All authors contributed equally.}}% <-this % stops a space
\IEEEtitleabstractindextext{%
\begin{abstract}
The protein-protein interaction (PPI) network provides an overview of the complex biological reactions vital to an organism's metabolism and survival. Even though in the past PPI network were compared across organisms in detail, there has not been large-scale research on how individual PPI networks reflect on the species relationships. In this study we aim to increase our understanding of the tree of life and taxonomy by gleaming information from the PPI networks. We successful created (1) a predictor of network statistics based on known traits of existing species in the phylogeny, and (2) a taxonomic classifier of organism using the known protein network statistics, whether experimentally determined or predicted \textit{de novo}. With the knowledge of protein interactions at its core, our two models effectively connects two field with widely diverging methodologies--the phylogeny and taxonomy of species.
\end{abstract}}

% \begin{IEEEkeywords}
% Computer Society, IEEEtran, journal, \LaTeX, paper, template.
% \end{IEEEkeywords}}

\maketitle

\IEEEdisplaynontitleabstractindextext

% \fi
\IEEEpeerreviewmaketitle

\IEEEraisesectionheading{\section{Introduction}\label{sec:introduction}}

\subsection{Motivation}
\IEEEPARstart{T}{he} wide availability of protein-protein interaction (PPI) networks across vast number of organisms provide an unique opportunity to increase understanding of the relationships between species. The taxonomic and phylogenetic trees describe the similarity in physical characteristics and evolutionary relationships between the species, respectively. Since the advent of molecular biology, new techniques emerged that can facilitate the traditionally classification of species based on morphological and fossil evidence, such as the molecular clock hypothesis \cite{thorpe1982molecular}, genomic phylostratigraphy \cite{domazet2007phylostratigraphy} and molecular phylogeny \cite{teeling2005molecular}.

% In this study, we aim to understand 1) how PPI networks in simpler life forms differ from those in more complex life forms, 2) how species in different phylogenetic groups exhibit different network characteristics, and 3) how closely related species have similar characteristics in their interaction networks.

Past studies indicated that the protein-protein interaction networks of species is a reflection of the evolution of both individual proteins and the organisms as a whole. This suggest that the information gleamed from the networks can be used to classify or predict species in relations with their evolutionary trajectories and taxonomic classification. This leads to our observation that the protein-protein interaction network is effectively a bridge between the phylogeny and taxonomy of species, and the conclusions made on protein networks can be used to gain insight or make prediction in both directions between phylogeny and taxonomy. In this study, we would use overall network statistics (like clustering coefficient) of PPI networks for different species.

\subsection{Problem Definition}
Briefly our work proceeds as follows: first, we explore how the various network statistics compare and differ throughout the tree of life and get deep insights into how protein interactions have evolved over time. Following this, given the position in the tree of life, we build a predictor for protein network statistics. Finally we create a taxonomic classifier which takes as input the protein network statistics (either experimentally determined or obtained \textit{de novo} with the predictor) and returns the predicted taxonomy of the species.

As the construction of PPI networks remains a painstaking task, typically biologists determine phylogeny and the taxonomic classification of a new species before the PPI network is created. Constructing PPI networks also requires huge amounts of financial and research resources. The potential of our model lies in the calculation of various protein network statistics before any protein interactions are determined, as long as the phylogeny of an organism is understood. With the help of our predictor, biologists can predict statistics about the PPI networks of new species by just knowing its phylogenetic position. These predicted PPI network statistics can aid in constructing the PPI networks or finding loopholes/mistakes in the constructed networks. The predicted values can also be used to further predict interesting properties of the species such as its taxonomy (a method which we developed in this work). 

We then study the relationship between PPI network characteristics and taxonomy of the species by building a species classifier that determines the taxonomy given the PPI network statistics of the species. The classifier can be used for systematic and automated taxonomy assignment based solely on protein network statistics and to check the taxonomy determined through other methods.

The combination of the predictor and the classifier can prove to be another interesting and high impact tool. Given the phylogenetic position of a species we can use our predictor to predict its network statistics; then using these predicted network statistics, our classifier can determine the taxonomy of the species thereby bridging phylogeny to taxonomy through PPI networks. Even though we use the PPI network data to build this predictor and classifier, we can find the taxonomy of a species given its phylogeny without knowing any information about its PPI networks. Such a tool can explore the interdependence between the logically distinct fields of Taxonomy and phylogeny and provide a bridge between the fields.

% Our study of network properties across phylogenetic and taxonomic relationships provides us a way to estimate the network properties of a new species based on its predicted position in the phylogenetic and taxonomic trees.

% At the conclusion of our project, we shall create a new species classification tool based on the network characteristics of PPI networks of individual species. Our approach is novel on the basis of organized comparison between protein networks in order to gain insight into species relationships. Past research in this area mostly focuses on the microscopic evolution of protein nodes in single species networks, whereas our method is unique in its focus on the comparison across species networks, and the attempt to elucidate signature features of the network regardless of the way each protein node arises during the course of evolution.

\subsection{Protein-protein Interaction (PPI) Networks and its Reflection on the Evolution of Life}
Protein-protein interactions (PPIs) are vital to the operation of all cellular functions. Proteins are chains of amino acids linked by peptide bonds--the combinations of which yield a vast array of molecules with countless shapes and sizes. Proteins act alone or together to perform chemical transformations; transmit signals; and provide structure/organization, transport, and recognition among many other cellular roles.
However, previous studies have been done on multiple aspects of the protein network evolution and its relationship with 
Information obtained from PPI network databases enables creation of interaction networks, and allows for systematic investigation of the complex biological activities within the cell. With the increase in research on PPIs and advancement in computational tools, interaction networks have been constructed and analyzed for thousands of species. 

\subsection{Making Sense of the Diversity of Life:\\Phylogeny and Taxonomy}
In this study we consider two vastly different classification schemes biologists employ in their study of species relationships: phylogeny and taxonomy.

A phylogenetic tree describes evolutionary relationships of a group of species, and a phylogenetic tree depicts the evolutionary relationships among organisms, while each branch points indicate when new species emerge from a common ancestor. Therefore, each node in the phylogenetic tree is an actual species and all non-leaf nodes are ancestors of one or more organisms. Phylogenetic trees are usually based on morphological or genetic homology and are build by comparing anatomical traits, genetic difference and by using molecular clocks.

Taxonomy, on the other hand, is the study of organisms with the goals of classifying living and extinct organisms according to a set of rules \cite{TaxonomyandPhylogeny}. The rules specify a hierarchy of groups, which form the non-leaf nodes of the tree, whereas the organisms are assigned to the tree leaves based on similarities and dissimilarities of their characteristics. Taxonomy is usually richly informed by phylogenetics, but remains a methodologically and logically distinct discipline.

Our understanding of species taxonomy is constantly updated based on the increasing understanding of evolutionary relationships between species. Advancements in molecular biology also aid in our study of species relationships, including the use of molecular clocks (rates of change in RNA or DNA sequences) and identification of orthologous proteins across the phylogeny. As a result in this study we are going to consider both the taxonomy and the phylogeny tree in our quest of a automatic classification of life, and compare the viability of our model under various goals and settings. The predictor and the classifier we build can act as a link between them through PPI networks.

\section{Related Work}
Major research directions related to our theme involves study of protein networks, phylogenetic tree and taxonomy. But,to our knowledge, this is the first study attempting to bridge the knowledge of taxonomy and phylogeny using protein networks. 

\subsection{Emergence of Protein Nodes in Networks}
Past research on protein network evolution has focused on single species network by observing how the protein nodes arise in the evolutionary timescale. This is done by assigning an evolutionary age to each protein node using techniques like phylogenetic stratigraphy \cite{domazet2007phylostratigraphy}. Research has shown that in the course of evolution, the protein nodes are added to the network in accordance with the preferential-attachment model \cite{eisenberg2003preferential, vazquez2003growing}. Further research provides details on protein nodes emergence mechanisms like gene duplication and gene loss \cite{hancock2005simple, pastor2003evolving} or the divergence model \cite{ispolatov2005duplication}. Even though these studies tend to focus only on a single species and so are of interest primarily in the biological context, they can in fact supplement and support our observations during our comparison across the phylogenetic tree using hypotheses formed from protein emergence models.

\subsection{Protein Network Structure Evolution}Phylogenetic trees are usually based on morphological or genetic homology.
Research on protein network evolution has also focused on the high level structure of protein networks. Some studies identify motifs and community structures that are commonly preserved in the evolution process\cite{wuchty2003evolutionary, milo2002network}, and others have looked at the high level structural graph statistics in terms of the entire graph \cite{berg2004structure}. The studies suggest that proteins evolved earlier are more likely to stay on as local clustering structures \cite{jeong2001lethality}, and that the overall statistics of the graph is with a power law dynamics with the evolution \cite{bu2003topological}. This suggest that our attempt to compare graph statistics is based on the valid assumption that graph characteristics is related the way that the interactions links arise, and since the rudimentary species have a less-evolved network we can expect the network statistics can also reflect the fact.

\subsection{Protein Network Alignment Informs the Phylogenetic Tree}
Protein network alignment methods \cite{liao2009isorankn, kelley2004pathblast, kalaev2009fast, kalaev2008networkblast} can inform our approach and suggests way to effective summarize the comparison of networks across species. These methods attempt to assign node-node relationships for homologous proteins in different species \cite{koyuturk2006pairwise, singh2007pairwise}, and report a similarity score \cite{zhang2005tm} that can suggest evolutionary relatedness between species. 

\subsection{Evolution of Protein Network Topology}
Our work comparing protein network across species is unique in its focus on the phylogenetic relationships of species and its relatedness with the protein networks themselves. The work in previous studies \cite{milo2004superfamilies, maslov2002specificity} is focused primarily on the comparison of protein networks, and how it relates to the evolutionary relationships we've come to understand between the species. To this end the methods have been applied to generate an alternative phylogenetic tree based solely on the topological characteristics of the protein networks \cite{Kuchaiev:2010cj} using tree construction methods such as the average distance algorithm applied to similarity score between all the PPI networks, without regards to individual gene/protein evolution. This complements our observation that the summary statistics of networks can provide information on the species and its evolution.

\begin{figure*}[h!]
\centering
\subfloat[Depiction of bacteria, archaea, and eukaryota domains in the phylogenetic tree]{\includegraphics[width=0.37\textwidth]{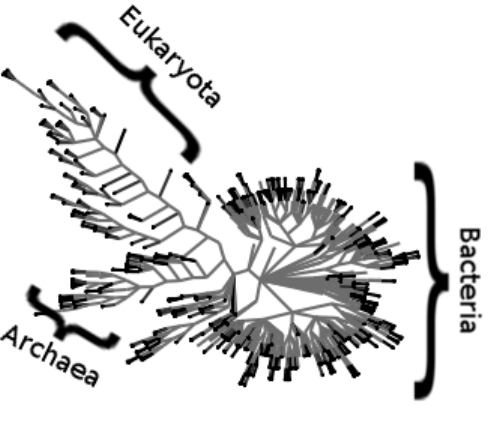}%
\label{tol:domains}}
\subfloat[Network diameter]{\includegraphics[width=0.37\textwidth]{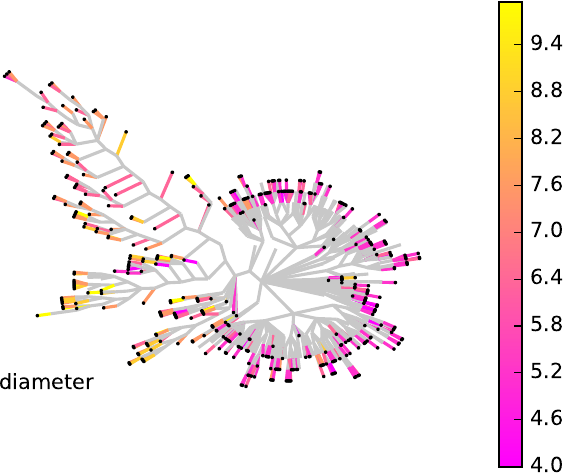}%
\label{tol:diameter}}
\hfil
\subfloat[Average degree]{\includegraphics[width=0.37\textwidth]{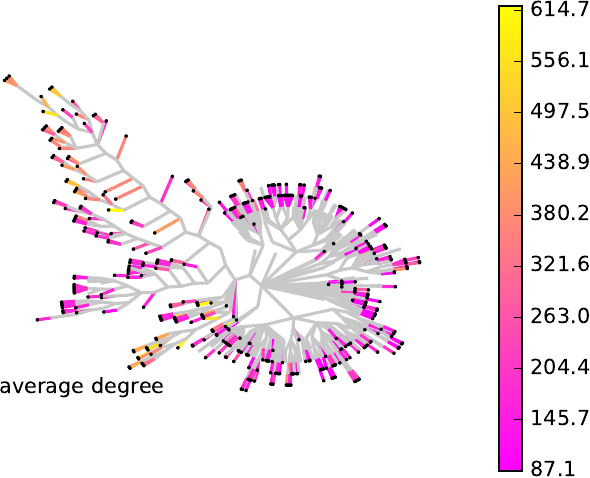}%
\label{tol:mean-degree}}
\subfloat[Standard deviation in degree distribution]{\includegraphics[width=0.37\textwidth]{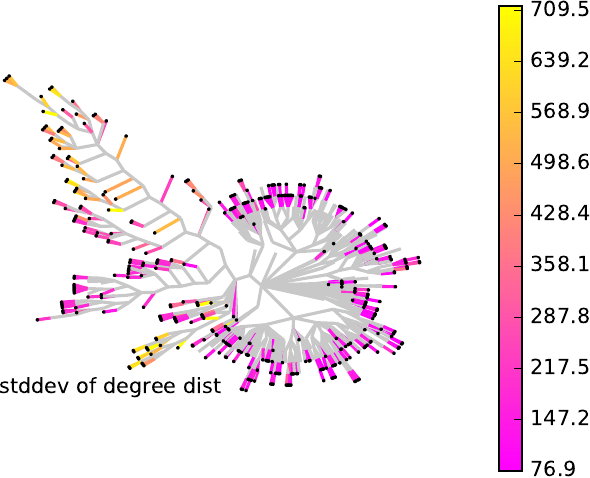}%
\label{tol:stddev-degree-dist}}
\hfil
\subfloat[Entropy of degree distribution]{\includegraphics[width=0.37\textwidth]{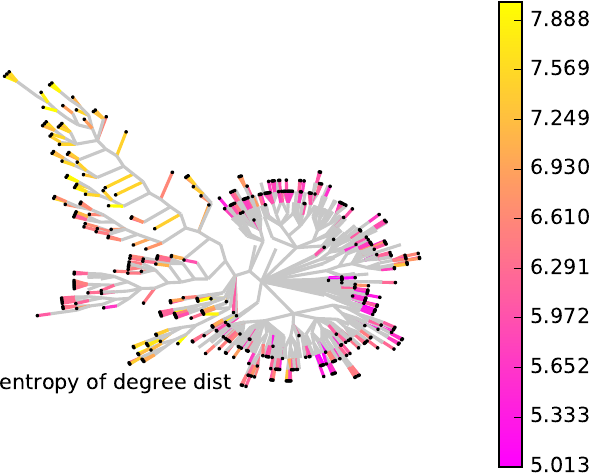}%
\label{tol:clustering-coeff}}
\subfloat[Spectral norm]{\includegraphics[width=0.37\textwidth]{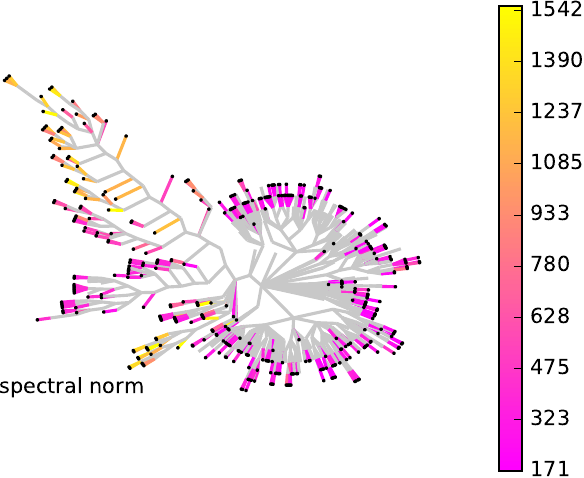}%
\label{tol:spectral-norm}}
% \subfloat[Fraction of nodes in the largest connected component]{\includegraphics[width=0.33\textwidth]{pdf/frac-connected-component-cropped}%
% \label{tol:frac-connected-component}}
\hfil
\caption{Heatmaps for spatially visualizing the variation of various network statistics over species in the phylogenetic tree of life. Only selected figures for network statistics which show interesting patterns are included for this report.}
\label{fig:tolheatmaps}
\end{figure*}

\section{Approach}
\subsection{Data collection}

\subsubsection*{Protein-protein Interaction Networks}
STRING v10 is the data source of choice for studies concerning PPI networks across multiple species, and version 10 of the database include PPI networks for more than 2000 species. There are multiple ways to define edges between the protein nodes, including functional relationships found with experimental means, known biological pathways, existing studies suggesting functional codependence, and \textit{de novo} computational prediction. 

We restrict our investigation to the more significant species within the dataset determined by the number of published work on the species when searching through PubMed. By limiting our work to species with more than 100 published work, we choose to focus on the 427 most studied organisms since they are of higher interest to biologists and have more complete protein networks. The problem remains, however, that the bias in network completeness (missing nodes in some of the species networks) might introduce artifacts in the result of our study. We therefore plan to take steps to ensure that the biases are not interpreted as part of our experimental conclusion.

Further, the data set also contained edges for protein protein interactions that were anticipated to be existing and not actually found in biological studies. These edges were removed as we wanted to extract features from the true network without any assumptions or predictions.

\subsubsection*{Phylogenetic Tree of Life}
In recent times, there has been ample work on the problem of constructing phylogenetic trees from molecular sequence data of species \cite{gouy2010seaview,shimodaira2001consel,van2009updated}. The STRING database itself provides a phylogenetic tree of species as accessory data to the interaction networks. Since the usage of identifiers is consistent across the interaction networks and the tree of species, it is a natural choice to work with the STRING tree of species. Figure \ref{tol:domains} shows the STRING tree of species annotated with the three domains (Archaea, Bacteria, and Eukaryota).

The STRING tree of species, however, does not provide any information on the similarity of species that are connected by an edge in the tree. To enrich the STRING tree of life with such information we use the tree of life developed in \cite{hug2016new}. Since the species identifiers used in \cite{hug2016new} are not consistent with those used in STRING, we use heuristics (based on lineage of species) to generate a mapping between the two sets of species identifiers.

\subsubsection*{Taxonomic Hierarchy}
The NCBI has made available the Taxonomy Browser tool \cite{federhen2012ncbi} which allows convenient browsing of genome information as well as taxonomic lineage of species. The species identifiers used in Taxonomy Browser are consistent with those used in STRING. Thus, to collect information about taxonomic hierarchy, we 
\begin{enumerate}
\item query the Taxonomy Browser with all the species in the STRING database,
\item scrape the lineage of species from the results HTML generated, and
\item build a directed tree by adding the lineages as paths in the tree.
\end{enumerate}

The final tree is further processed (as described in section \ref{sec:classifier}), before being used for construction of our taxonomic lineage classifier.

\begin{figure*}[h!]
\centering
\subfloat[Average degree]{\includegraphics[width=0.45\textwidth]{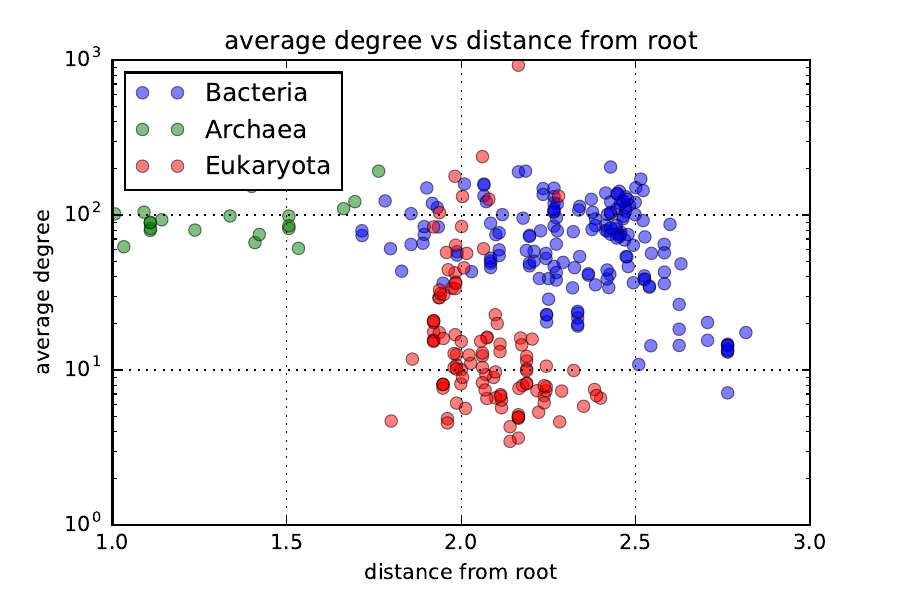}%
\label{fig:averagedegree}}
\subfloat[Density]{\includegraphics[width=0.45\textwidth]{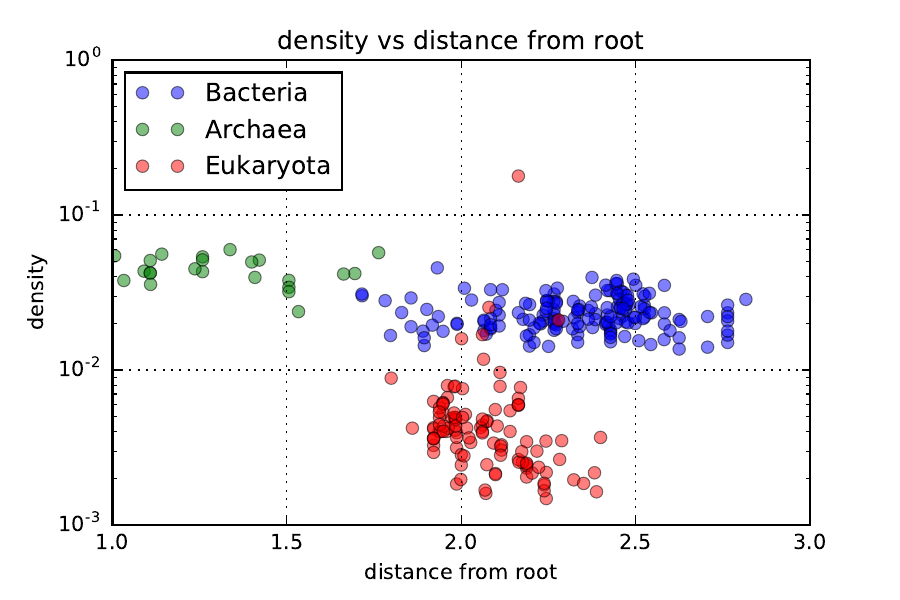}%
\label{fig:density}}
\hfil
\subfloat[Diameter]{\includegraphics[width=0.45\textwidth]{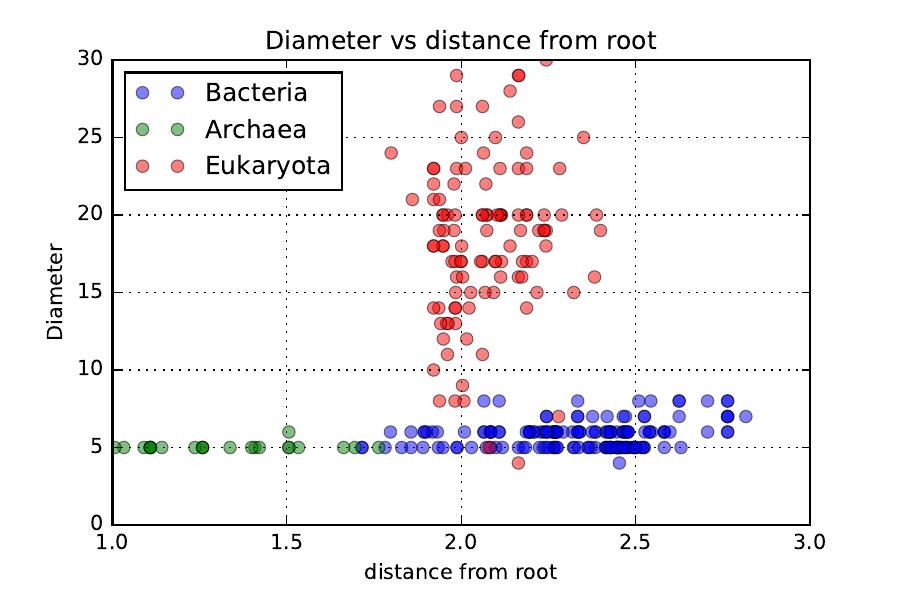}%
\label{fig:diameter}}
\subfloat[Global clustering coefficient]{\includegraphics[width=0.45\textwidth]{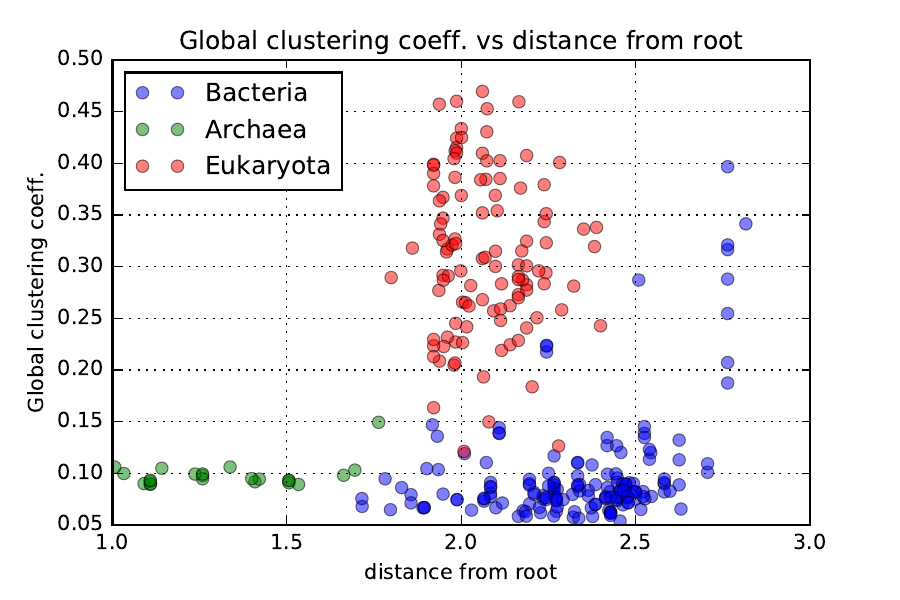}%
\label{fig:globalclusteringcoeff}}
\hfil
\subfloat[2-star density]{\includegraphics[width=0.45\textwidth]{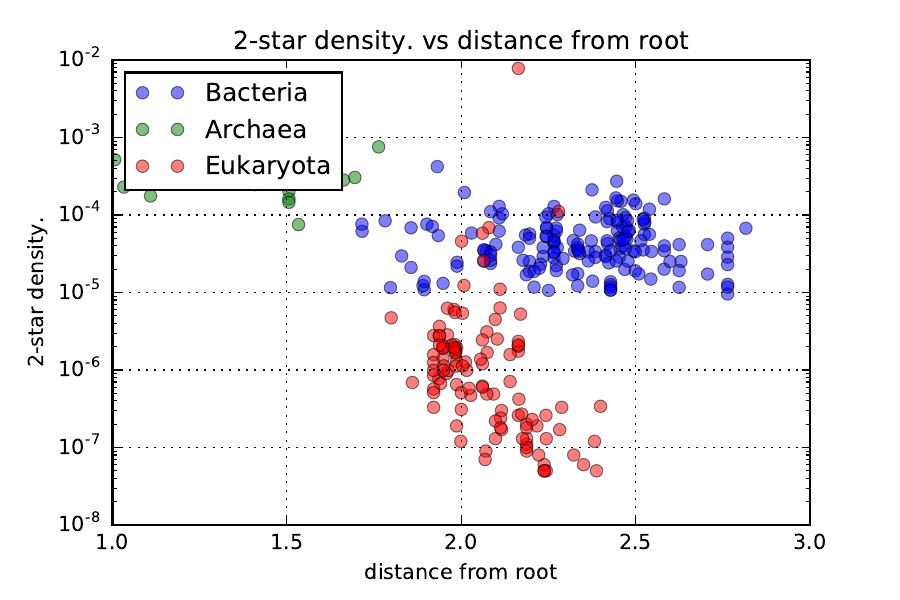}%
\label{fig:2stardensity}}
\subfloat[Edge entropy]{\includegraphics[width=0.45\textwidth]{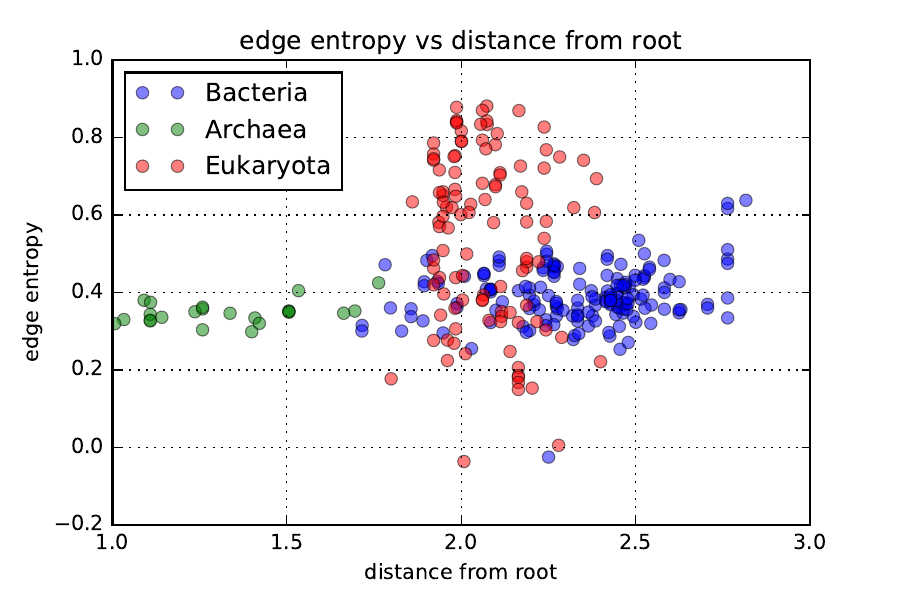}%
\label{fig:edgeentropy}}
\hfil
\caption{Variation of select network statistics with distance of species from the root in the phylogenetic tree of life. }
\label{fig:s}
\end{figure*}

\subsection{Network Statistics}
In the following section, we give a brief summary for each of the PPI network metrics that form the basis of our study, and are included as features in our classifier and predictor. With some exploratory work we found that the following set of features gave the best results.

\begin{enumerate}
	\item \textbf{Number of nodes}: This captures the number of protiens in the species and in turn the complexity of the species.
    \item \textbf{Number of edges}: This captures the number of types of protein-protein interactions, again indicating the complexity of the species. 
    \item \textbf{Average degree}: On an average, how many other proteins does a protein in this species interact with.
    \item \textbf{Maximum degree}: How many proteins does the most active protein in this species interacts with. This refers to the protein with the least specificity.
    \item \textbf{Density}: This is the ratio of number of edges to the number of possible edges in the network. This indicates the average specificity of proteins.
    \item \textbf{Number of connected components}: This captures groups of proteins that donot interact with one another.
    \item \textbf{Fraction of nodes in largest connected component}: These capture the complexity of biochemical reaction in species.
     \item \textbf{Full diameter}: Diameter is the longest of all shortest paths in the network. 
    \item \textbf{Global clustering measure}: It is ratio of number of closed triads to sum of closed and open triads.
    \item \textbf{Clustering coefficient}: The clustering coefficient of a node is defined as the fraction of existing edges among all the possible edges between the neighbor nodes. 
    \item \textbf{2-star density}: Ratio of number of 2-stars present to number of 2-stars possible in the network.
    \item \textbf{3-star density}:  Ratio of number of 3-stars present to the number of 3-stars possible in the network.
    \item \textbf{Entropy and Gini coefficient of degree distribution}: Multiple studies focusing on single species PPI network have shown that proteins emerge in accordance with the preferential attachment model \cite{eisenberg2003preferential,vazquez2003growing,vazquez2002modeling}. Even though in our study of protein networks we compare degrees between network, not protein nodes, we can expect the same properties of network growth in preferential attachment. We're interested in seeing how the average degree grows with respect to the evolutionary years and advancement in evolution. To compare degree distributions across networks, we collect meta statistics over the distribution like mean, gini coefficient, entropy, etc.
    \item \textbf{Assortative mixing} : This captures preference for a node to be attached with a node thats similar to it in terms of degree.
    \item \textbf{k-cores features}:  A k-core of a graph is a maximal connected subgraph of the graph in which all vertices have degree at least k. We compute the maximum k, such that k-core exists and use that k, number of nodes in that subgraph, number of edges in that subgraph as features. 

\end{enumerate}

We believe the above features would rightly reflect the complexity of protein-protein interactions and other biochemical pathways of different species.

\section{Analyzing the Dataset}
\label{sec:analyzing}
We devoted a large amount of effort on the exploratory analyses of the dataset, and we were able to report some of the important observations we gathered. We focus on the following analysis in this section:
\begin{enumerate}
\item \textbf{Spatial visualization of variation of network features across the phylogenetic tree: }to help us identify macro-level relationships and form hypothesis about observed patterns.
% \item \textbf{Correlation of network features with amount of published work: }in order to understand whether a particular pattern in the features is an artifact of the relative amount of completeness (or incompleteness) of networks.
\item \textbf{Variation of network features at different stages of evolution: }we aggregate the network features for all species at the same number of branch hops from the root in the phylogenetic tree, and report the variation of these aggregates with the number of branch hops.
\end{enumerate}

\subsection{Visualization of network features on the Phylogenetic tree}
\label{sec:visualizingtol}

To show how the various network metrics varies throughout the entire phylogenetic tree, we generate heat maps of the phylogenetic tree with each point showing the magnitude of the various network characteristics (see Fig. 1). 

We observe that there are obvious differences between species in the Eukaryota domain and the other prokaryotes (Archaea and Bacteria). This suggest that we should achieve at least moderate success in our effort to classify organism based on PPI network statistics. 

While we omit a statistical evaluation of the hypothesis that network properties are different across different domains of species, we outline such a procedure. A simple null model for comparison would be the phylogenetic tree with its node labels (and corresponding values of network statistics) shuffled randomly. The differences in size and distribution of clusters (computed by binning the network statistic values into small number of bins and using any of the spectral clustering algorithms) shall give us a meaningful quantification of our hypothesis.

% \subsection{Correlation of network features with amount of published work}
% As previously noted, not all species have received equal attention with regards to the study of their protein-protein interaction. As a result, the PPI networks of certain (groups of) species may be incomplete as compared to those of species (such as human, monkey, mice etc.) that have been studied at a greater extent. 

% In order to understand whether the observed patterns in network characteristics are merely artifacts of incompleteness (or completeness) in networks, we report the Pearson correlation and the corresponding p-values (see Table 1) for identifying the correlation between amount of published work and the various network statistics computed. We use the number of PubMed articles that mention a particular species as a proxy for the completeness of the species' PPI network.

% From the r and p-values computed, we infer that the following network statistics are correlated with amount of published work: 
% \begin{enumerate}
% \item Average degree
% \item Standard deviation in degree distribution
% \item Entropy of degree distribution
% \item Spectral norm
% \end{enumerate}

% Therefore, in our further analysis, we shall restrict the use of the above statistics. 

\subsection{Variation of network features with degree of evolution}
\label{sec:networkfeaturesevolution}
The tree of life with edge weights obtained from \cite{hug2016new}, gives us a measure of dissimilarity of all pairs of species that are connected by an edge. Since the phylogenetic tree encodes evolutionary information such as ancestry and common descent, we use the tree distance (sum of weights of edges on a path) of a species from the root of the phylogenetic tree as a proxy for its degree of evolution. It is observed that Bacteria, in general, are further away from the root than Eukaryotes and Archaea--which can be reasoned since Bacteria were among the very first living organisms to appear on Earth. 

In search of interesting patterns, we studied the variation of network statistics of species with the distance from the phylogenetic root of that species as described above. A selection of interesting results is presented in Figure \ref{fig:s}. It can be seen from the figures that Archaea are species with lowest distance from the phylogenetic root, and Bacteria are located farther away from the root along with Eukaryota.

While we did not observe a simple dependence of any network property on the distance from root: we remark that the average degree, density, 2-star density tend to decrease as species move farther away from the root, whereas there is a weak opposite trend for other statistics such as diameter, global clustering coefficient, and edge entropy. Further, in all the included figures, it can be seen that there is a rather apparent distinction between different domains based on various network statistics. For Bacteria and Archaea in particular, we can see the variation of network statistics within the domain is relatively small. These observations lead us to conclude with confidence that it is possible to classify species into taxonomic groups (at least at the level of domains) using only the network statistics of their PPI networks.

\section{Predicting Network Statistics from Position of Species in Phylogenetic Tree}
\label{sec:predictingstatistics}
As we identified statistics that are meaningful representation of the protein networks, our next step is to build a predictor of the same statistics based on features gleamed based on the phylogeny of species. We construct and test our model by splitting the 427 species we included into the train set and the test set with a random 80/20 split. Network statistics of the nodes in the train set is presumed to be known, whereas statistics of the test set are those we set out to predict. The prediction is based on the known position of species in both the train and test sets on the phylogenetic tree of life. Position refers to the exact location of the species in the tree - the parent and the children can precisely define it. Lets look at the features that can capture the position of a node in the tree. 

\subsection{Feature Extraction}
\subsubsection{Sibling Feature}
To calculate the sibling features, for every species we find the latest ancestor that contains more than one descendants in the tree; these descendants are considered the "siblings" of the species node in question. We then take the mean of all the network statistics of these siblings to generate a set of sibling features that can be used to predict the network statistics of the particular node. 

\[feature_{{\mathop{\rm sib}\nolimits} }^{{\mathop{\rm stats}\nolimits} } = \frac{{\sum\limits_{{\mathop{\rm siblings}\nolimits} } {stats} }}{{\left| {{\mathop{\rm siblings}\nolimits} } \right| }}\textrm{   for every } stats\] 

% \subsubsection{Parent Feature}
% Since most ancestors of the species nodes are internal nodes which lack the PPI network data in the database, we estimate the network statistics of the parent by averaging over all descendants of the latest ancestor with more than one descendants. To improve the estimation, we include the statistics of all the 2417 protein networks when finding the descendants of the species node in question, instead of using only the 427 species considered in this study (like we did with the sibling features). Therefore for each of the 427 species nodes, we obtain a set of parent features by averaging across all descendants in the 2417 nodes.

% \[feature_{{\mathop{\rm par}\nolimits} }^{{\mathop{\rm stats}\nolimits} } = \frac{{\sum\limits_{{\mathop{\rm descendants}\nolimits} } {stats} }}{{\left| {{\mathop{\rm descendants}\nolimits} } \right| }}\textrm{   for every } stats\] 

\subsubsection{Cousin Feature}
Cousins of a node are defined as the nodes with the same level from root in the tree as the node under consideration. To get the cousin features, we look at the entire phylogeny to find in the 427 species those that are of the same hops from root as the species in question. By doing this we can have a estimation of how the hops from root affects network statistics of a particular species we set out to predict.

\[feature_{{\mathop{\rm cuz}\nolimits} }^{{\mathop{\rm stats}\nolimits} } = \frac{{\sum\limits_{{\mathop{\rm cousins}\nolimits} } {stats} }}{{\left| {{\mathop{\rm cousins}\nolimits} } \right| }}\textrm{   for every } stats\] 

% \subsection{ Construction \& Evaluation}
% Using the extracted features, we then attempt to predict network statistics based on the features obtained from the species' position in the phylogeny. We trained our support vector machine model on the siblings, parent and cousins feature obtained from species 

\subsubsection{Caveats in Feature Extraction}
During feature extraction, our first intuition is to create features based solely on the species node position in the phylogeny instead of taking the network statistics into account. This is motivated by the fact that a lot of the internal nodes in the phylogeny do not include protein networks data, so it's impossible to obtain network statistics of these nodes without estimation. In our attempt to create features that summarize the position of species nodes in the tree of life, we created a bit vector for every internal node in the phylogeny indicating which of the 427 species are descendants of the internal node. We found as a result that the feature vectors are too large and too sparse (since we include all the internal nodes as possible ancestors) and perform poorly even with dimensionality reduction using principal component analysis (PCA). 

We therefore conclude that in order to effectively predict the network statistics of species in the test set we need to tap into the network metrics instead or relying merely on ancestry in the phylogeny. This is meaningful in the suggestion that the phylogeny itself is insufficient in accurately determining the traits of protein networks, and shows that the network nodes are highly influenced by both the evolutionary relationships and the traits of existing network statistics of nearby species in the phylogenetic tree.

\subsection{Predictor Model Evaluation}

We first divide the species into a train and test set with a random 80/20 split. From every network statistics, we train a linear support vector machine (SVM, linear kernel with C = 100) model based on the sibling and cousin features of the species in the train set. We then predict each of the various network statistics for species in the test set using the , and then compare the predicted values with the actual network statistics using the relative error:

\[{\textrm{relative error}} = \frac{{|predicted - actual|}}{{actual}}\]

The relative errors are then summarized across all species in the test set, and for each type of network statistics we obtain a mean and standard deviation of the particular statistics (Table \ref{table:predictorresults}).

\begin{table*}
\centering
\normalsize
\caption{The mean and standard deviation of the relative error of the predicted neted values are obtained with SVM models trained using features from the sibling and cousin nodes of species in the train set, and tested using the same features obtained with species in the test set.}
\begin{tabular}{||l|r|r||}
\label{table:predictorresults}
{} &  Mean of Relative Error &  Standard Deviation of Relative Error \\ \hline
Nodes                            &         0.244 &         0.233 \\
Edges                            &         0.739 &         0.978 \\
Average Degree                   &         0.371 &         0.560 \\
Maximum Degree                   &         0.381 &         0.349 \\
Density                          &        13.223 &        14.857 \\
Number of Components             &         1.771 &         7.575 \\
Maximum Component Size           &         0.123 &         0.340 \\
Full Diameter                    &         0.128 &         0.111 \\
Effective Diameter               &         0.106 &         0.181 \\
Global Clustering                &         0.405 &         0.395 \\
Clustering Coefficient           &         0.136 &         0.103 \\
Star Density 2                   &         1.366 &         2.494 \\
Star Density 3                   &        14.120 &        41.749 \\
Gini Coefficient of Degree Dist. &         0.089 &         0.061 \\
Edge Entropy                     &         0.025 &         0.014 \\
Assortative Mixing               &         0.186 &         0.235 \\
K-cores Maximum Degree           &         0.337 &         0.364 \\
K-cores Maximum Nodes            &         0.664 &         0.796 \\
K-cores Maximum Edges            &         2.293 &         4.234 \\
\end{tabular}
\end{table*}

\subsection{Analysis}
With results from Table \ref{table:predictorresults} we can conclude that the predictor works well with most network statistics considered in this study. The mean error is in the range of 10-20\% for most of the network statistics, and the standard deviation is also considerably low. Interestingly, it seems that a few of the network statistics are extremely hard to predict, like the density and star density of the networks. 

The reason that this model performs so poorly in a few cases suggests that some of network statistics are quasi-random and possibly unrelated to the evolutionary relationships. Our observation was that the star density and the density of a protein-protein interaction network does not have readily available biological interpretation; in other words, they are irrelevant in the context of protein interactions and biochemical reactions in an organism.

The results we obtained can also inform our understanding of the interaction between evolution and network growth. As we can see in the Table 2, the diameter of protein network, the maximum component size and a few variance measures of the degree distribution performs remarkably well in our model; this combined with the biological interpretation detailed in Section 3.2 suggests that the graph statistics are readily interpretable in the context of evolutionary biology.

Noting that statistics like number of edges are huge numbers and are generally orders of magnitude different for different species, we believe predicting the number of edges with 70\% accuracy is still good result considering the underlying nature of these statistics. Also, we can derive the number of edges with a better accuracy by using predicted average degree and number of nodes as they have far lower error rates.

\section{Taxonomic Lineage Classifier}
\label{sec:classifier}

\begin{figure}[h!]
\label{fig:taxonomichierarchy}
\includegraphics[width=\linewidth]{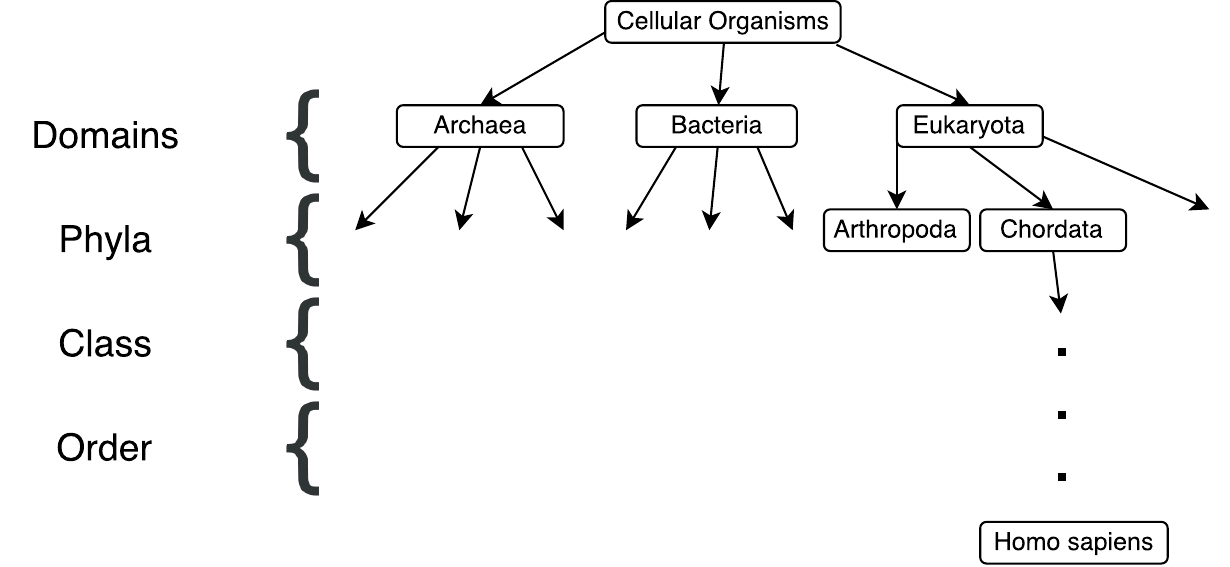}
\caption{A partial taxonomic hierarchy shown as a rooted directed tree}
\end{figure}
% praty writing this section

% intro about the task
One of the glaring observations from the analysis in Sections \ref{sec:visualizingtol}-\ref{sec:networkfeaturesevolution} was that PPI network statistics differ significantly across species in the Archaea, Bacteria, and Eukaryota domains. It is apparent that given the PPI network statistics of a species, it is possible to classify it with reasonable accuracy into one of the three domains. While in the interest of brevity, we could not include a similar analysis which compares network statistics across finer levels of taxonomic hierarchy, we observe that many network statistics tend to be similar for species in the same taxonomic group and different for species across different groups. Naturally, we were tempted to ask ``Can a species'  PPI network statistics be used to predict its taxonomic classification?''

It should be noted that the taxonomic classification of a species and the construction of its interaction species are two tasks that are fairly independent of each other. Our work aims to form a bridge between the the science of taxonomy and molecular interaction biology. Further, if the taxonomic lineage classifier is coupled with the network statistics predictor (discussed in Section \ref{sec:predictingstatistics}), it will be possible for biologists to discover the taxonomic lineage of a newly discovered species using only it's hypothesized position in the phylogenetic tree.

\begin{figure}
\label{fig:rfeclassifier}
\includegraphics[width=\linewidth]{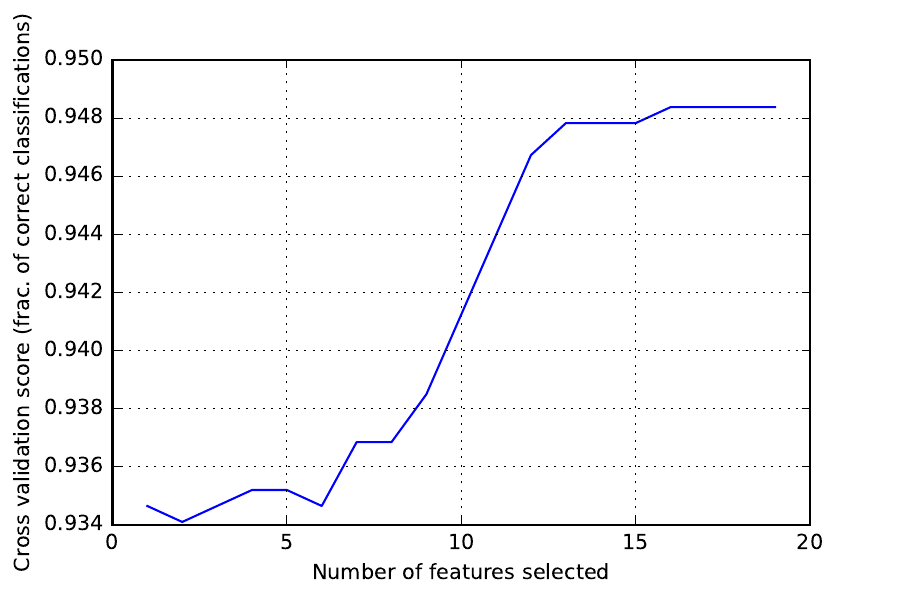}
\caption{Performing recursive feature elimination on the classifier for domains shows us that highest accuracy is achieved when using all of the 19 features.}
\end{figure}
\subsection{Approach}
The taxonomic hierarchy is a rooted directed tree, where internal nodes represent taxonomic groups and the leaves represent individual species (see Figure \ref{fig:taxonomichierarchy}). Each path from the root to a leaf (species) represents the taxonomic lineage of that species. Thus, the classification task requires us to classify a species (using its PPI network statistics) into one of such paths.

We approach the lineage classification task as a hierarchy of classification tasks, each of which classifies a species into finer taxonomic groups given that it has been correctly classified to some coarser taxonomic group. In other words, we construct one classifier for each internal node in the taxonomic hierarchy (tree), whose role is to classify an example into a taxonomic group represented by one of the children of this node.

To construct the full taxonomic lineage of an example (network statistics of a species), we use the classifiers at successive levels as follows:
$$ c_i = \text{classifier}_{c_{i-1}}(x)$$
$$ c_0 = \text{cellular organisms (root)}$$
where $x$ is the feature vector (of network statistics) to be classified, $\text{classifier}_{c_{i-1}}$ is the classifier corresponding to node $c_{i-1}$ in the taxonomic hierarchy, and $c_i$ represents a node in the taxonomic hierarchy at distance $i$ from the root. The lineage of the species with feature vector $x$ is then $(c_0, c_1, \dots, c_{n-1})$ where $n$ is the height of the taxonomic hierarchy.

Each of the classifiers is a SVM with linear kernel $(C=100)$. $\text{classifier}_{c_i}$ is trained on species for which $i$-th level in lineage equals $c_i$.

The set of features used for classification is the same as those computed by the predictor in Section \ref{sec:predictingstatistics} (see Table 2). Since the linear kernel SVM allows us to extract weights corresponding to each feature, we performed recursive feature elimination to minimize the number of features used. In Figure \ref{fig:rfeclassifier}, it can be seen that it is best to use all of the 19 network statistics as features.

\subsection{Caveats}
The taxonomic hierarchy described in the previous sections is not exactly well-structured as described. In particular,
\begin{enumerate}
\item The length of lineage is different for different species, i.e. the height of the taxonomic hierarchy is not uniform.
\item The levels in the lineage do not necessarily map to the taxonomic levels of domain, kingdom, phylum, class, order, family, genus, species.
\item For many intermediate levels of lineage that do not correspond to any of the above taxonomic levels, there is just one child node in the taxonomic hierarchy.
\end{enumerate}
For the purposes of this project, we restrict our dataset to only those species for which we can extract information for the taxonomic levels of domain, kingdom, phylum, class, order, family information from the lineage. We also collapse all the intermediate levels in the lineage which do not correspond to any of the above taxonomic levels. Since there are not enough training examples within each genus, we restrict our classification to the above six taxonomic levels.

\begin{figure}[h!]
\includegraphics[width=\linewidth]{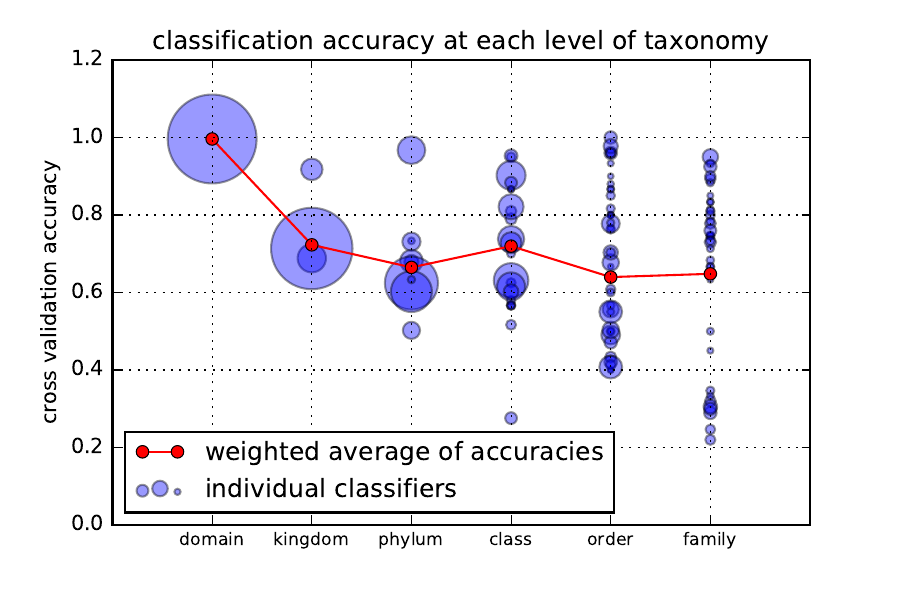}
\caption{Each blue disc marks the cross-validation accuracy of a particular classifier whose taxonomic level is read from the X-axis. The area of the disc is proportional to the number of training examples that were used for that classifier. The weighted average of accuracies is plotted in red.}
\label{fig:classifierresults1}
\end{figure}

\begin{figure}[h!]
\includegraphics[width=\linewidth]{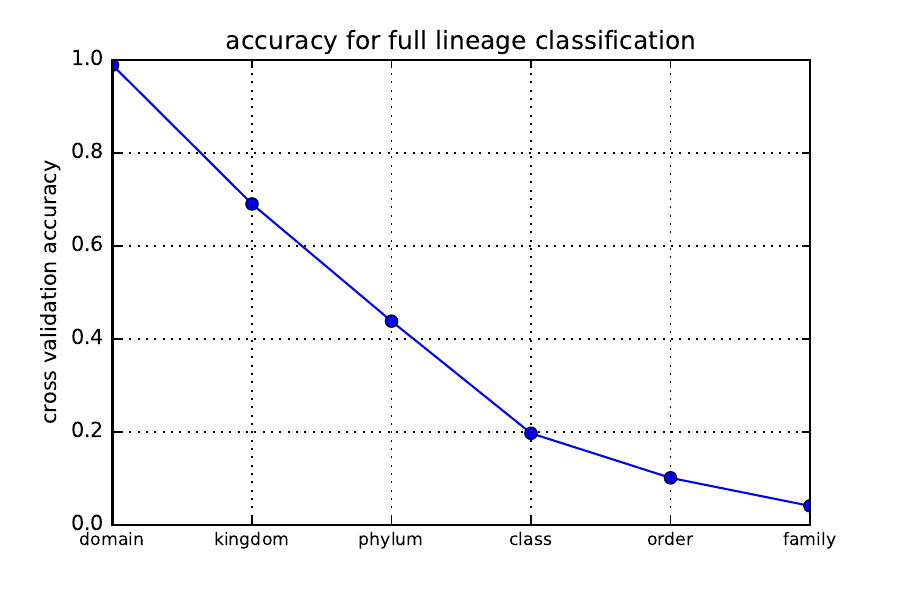} 
\caption{In contrast to Figure \ref{fig:classifierresults1}, here we show the accuracy for the full lineage classification. Almost 100\% accuracy is achieved at the domain level, whereas lower than 10\% accuracy is achieved for the lineage classification till the taxonomic level of family.}
\label{fig:classifierresults2}
\end{figure}

\subsection{Results}
We perform a five-fold cross validation on the dataset reduced to 1537 species after filtering. Note that we also include the species with lower publication counts for this particular problem. 

The cross-validation accuracies for each of the individual classifiers (one for each node in the taxonomic hierarchy) are shown in Figure \ref{fig:classifierresults1}. In the figure, each blue disc represents the accuracy of a single classifier. The plot is arranged in order of taxonomic levels and a tree of classifiers can be observed from the figure. We note that the classifier for discovering the domain of a species performs exceptionally well. While the classifiers at lower levels of taxonomy perform relatively poor, but considering that each of them have much lower amount of training data available and conceivably larger number of labels to classify into, their performance seems to exceed our expectations. 

The cross-validation accuracy for the full lineage classification is shown in Figure \ref{fig:classifierresults2}. The plot shows the classification accuracy up till various taxonomic levels. It can be seen that the classifier (rather, a hierarchy thereof) performance gets poorer as we seek to discover finer taxonomic classifications for a species. It is worth mentioning that for a test set of 365 species, our hierarchy of classifiers was able to predict the full lineage correctly for 15 species,  while predicting the domain correctly for 361 species.

\subsection{Analysis}
The results, particularly from Figure \ref{fig:classifierresults2}, lead us to conclude that our hierarchy of classifiers can be used to predict the lineage (at least partially) of a species using only the network statistics of its PPI networks. This is particularly impressive considering the fact that the taxonomy is constructed not merely with the evolutionary relationships of species, but rather based on a set of rules on the physical, physiological and other aspects of species traits.

With our varying degree of success applying the classification model to the different granularity of the taxonomic hierarchy, we can identify classification rules that are more relevant to the inherent biological process of the species. In addition to suggesting the taxonomy of newly discovered species, the same model can be apply even to organisms that are already classified, and suggest alternative taxonomy when building a newer model of classification. The classifier can also be used to make sense of the taxonomic rules, and can inform proposed modifications to the rules so that they are more relevant to the biological pathways, and complement existing molecular biology techniques in terms of species classification.

\section{Conclusion}

% In this part of the project, we have extensively gathered the required network statistics from our species of interest and plotted the stats with evolution and with respect to taxonomies. We have also found features which are not biased by the variation in research that has taken place with different species. We have also found some interesting patterns and their biological implications.

% In the next part of the project, we plan to frame more biological hypothesis based on our statistics. Now that we have recognized features which are not correlated with the bias in research, we will use these features to build a classifier, where, given the PPI network, we build a classifier to find out its taxonomy and position in the tree of life. Such an system will be useful to verify the placement of species in the tree of life.

% We would also be using the features we got till now to build a regression engine. Where given the position in the tree, we predict the new species's PPI network properties. When biologists find new species, they first find the position in the tree but need to do a lot of research to get PPI network statistics. it will be useful to build a system to predict PPI network statistics, given the position in the tree of life. We would be using hop from root, features of parent, aggregate features of cousin nodes, etc to build this regression model.

% Some species will be removed and will be used as a test set to evaluate our model.

With our phylogenetic predictor, we are able to provide an automated model that predict graph characteristics \textit{de novo}. The model is unique in the way it summarizes the statistics based solely on the characteristic of neighboring nodes on the phylogeny, and can generate with reasonable high accuracy the graph traits. As the construction of PPI networks for a new species is a painstaking process and generally occur long after the species is well studies in both evolutionary relationships and physiological traits, the predictor model can serve as a useful informant for biologists both during the discovery of newer species in cases where the PPI network is still incomplete for an organism.

With the taxonomic classifier we constructed based on characteristics of the protein networks of species, we are able to suggest taxonomic hierarchy of an existing species with varying level of success in the different levels. This is an interesting conclusion considering that the the taxonomy is determined generally unrelated to the physiology and internal biochemical reactions of species. The varying level of success, especially on the more granular level, suggest alternative taxonomy can be in place based on the discoveries enabled by our model. In addition, when a biologist can determine that network statistics of an organism with a level of certainly, the classifier can automatically suggest the taxonomic hierarchy of effectively especially on the higher level.

With our work on both the phylogeny and taxonomy of biology, we are able to connect the two fields through the bridge of the protein-protein interaction networks. Even though the fields of taxonomy and phylogeny are similar in their goal of making sense of the diversity of life, they employ vastly different approaches and often draw contrasting conclusions on the relationships between organisms. The fact that our two model connect the two fields makes possible a all-encompassing predictor that predicts the taxonomy of a species based on the known evolutionary relationships, even without prior knowledge of the protein-protein interaction networks. This conclusion comes to suggest an inherent relatedness between the fields of study, and we look forward to exploring the success of our model at different taxonomic levels. We also plan to engage in exploration in the other direction, going from the known taxonomy of a species and try to deduce its phylogenetic relationships towards others. Future work can also include coming up with better features for representing position in the phylogeny tree, using deep learning models for learning and investigating methods to improve accuracy of the classifier at  lower levels of taxonomy despite less training data. 

\ifCLASSOPTIONcaptionsoff
  \newpage
\fi

\bibliographystyle{IEEEtran}
% % argument is your BibTeX string definitions and bibliography database(s)
\bibliography{bare_jrnl_compsoc}

\end{document}